\documentclass{article}

\usepackage{PRIMEarxiv}
\pdfoutput=1
\usepackage[utf8]{inputenc} 
\usepackage[T1]{fontenc}    
\usepackage{hyperref}       
\usepackage{url}            
\usepackage{booktabs}       
\usepackage{amsfonts}       
\usepackage{nicefrac}       
\usepackage{microtype}      
\usepackage{lipsum}
\usepackage{caption}
\usepackage{subcaption}
\usepackage{fancyhdr}       
\usepackage{graphicx}       
\graphicspath{{media/}}     
\usepackage[affil-it]{authblk}

\usepackage[
backend=biber,
style=nature,
]{biblatex}

\bibliography{references}  
\usepackage{xcolor} 
\usepackage{setspace}

\title{Augmenting medical image classifiers with synthetic data from latent diffusion models
}

\author[1]{Luke W. Sagers\protect\footnotemark[1]\hspace{0.5em}}
\author[1]{James A. Diao\protect\footnotemark[1]\hspace{0.5em}}
\author[1,2]{Luke Melas-Kyriazi\protect\footnotemark[1]\hspace{0.5em}}
\author[3]{Matthew Groh}
\author[1]{Pranav Rajpurkar}
\author[4]{Adewole S. Adamson}
\author[5]{Veronica Rotemberg}
\author[6,7]{Roxana Daneshjou}
\author[1]{Arjun K. Manrai\protect\footnotemark[2]\hspace{0.5em}}

\affil[1]{Department of Biomedical Informatics, Harvard Medical School, Boston, MA, USA}
\affil[2]{Department of Engineering Science, University of Oxford, Oxford, UK}
\affil[3]{Kellogg School of Management, Northwestern University, Evanston, IL, USA}
\affil[4]{Division of Dermatology, Dell Medical School, The University of Texas at Austin, Austin, Texas, USA}
\affil[5]{Dermatology Service, Memorial Sloan Kettering Cancer Center, New York, NY, USA}
\affil[6]{Department of Dermatology, Stanford School of Medicine, Redwood City, CA, USA}
\affil[7]{Department of Biomedical Data Science, Stanford School of Medicine, Stanford, CA, USA}

\renewcommand{\thefootnote}{\fnsymbol{footnote}} 

\begin{document}
\maketitle

\footnotetext[1]{Equal contribution} 
\footnotetext[2]{Correspondence to Arjun\_Manrai@hms.harvard.edu}

\renewcommand{\thefootnote}{\arabic{footnote}} 

\begin{abstract}
While hundreds of artificial intelligence (AI) algorithms are now approved or cleared by the US Food and Drugs Administration (FDA)\cite{Wu2021-wk}, many studies have shown inconsistent generalization or latent bias, particularly for underrepresented populations\cite{Obermeyer2019-nu, Rajpurkar2022-xf, Mehrabi2021-ky}. Some have proposed that generative AI\cite{Ramesh2021-in,Rombach2021-vh} could reduce the need for real data\cite{Burlina2019-rb}, but its utility in model development remains unclear. Skin disease serves as a useful case study in synthetic image generation due to the diversity of disease appearance, particularly across the protected attribute of skin tone. 
Here we show that latent diffusion models can scalably generate images of skin disease and that augmenting model training with these data improves performance in data-limited settings. These performance gains saturate at synthetic-to-real image ratios above 10:1 and are substantially smaller than the gains obtained from adding real images. 
As part of our analysis, we generate and analyze a new dataset of 458,920 synthetic images produced using several generation strategies.
Our results suggest that synthetic data could serve as a force-multiplier for model development, but the collection of diverse real-world data remains the most important step to improve medical AI algorithms.
\end{abstract}

\keywords{Generative AI \and Medical Imaging \and Diffusion Models}

\section{Introduction}
Image-based machine learning algorithms are increasingly deployed in healthcare settings and directly to patients\cite{Wu2021-wk,Esteva2017-sb, Liu2020-ya, Balasubramanian2023-wm}.  As these algorithms enter real world use, concerns persist regarding representation of rare diseases and historically underrepresented groups\cite{Groh2021-zx, Daneshjou2021-gp, Daneshjou2022-vk, Kinyanjui2020-tn}. Augmenting model training with synthetic data from generative AI has been proposed as a tool for achieving desired performance and fairness objectives\cite{Ktena2023-sl, Rajotte2022-pi, Azizi2023-wr, Saharia2022-gc, Rajotte2021-pa}. Recent breakthroughs in latent diffusion models\cite{Ramesh2022-xz, Saharia2022-gc} have enabled generation of photorealistic images conditioned on text alone, images alone, or text and images in combination, with compelling results in dermatology applications\cite{Ktena2023-sl, Sagers2022-nn}. Some have suggested that synthetic data may obviate the need for additional data collection or substitute completely for real datasets\cite{Burlina2019-rb}. This view is controversial given that reported results have been inconsistent, that primary causes of performance improvements are not well-characterized, and that synthetic data could itself contain hallucinations or encode bias.

To understand the potential opportunities and pitfalls of synthetic data, our study isolates several potential sources of performance gains arising from augmenting deep learning model training using synthetic images.  In particular, we focus on algorithms for classifying skin disease, since the diversity of skin pathologies and skin tones provide an informative use case for probing issues of fairness and generalizability\cite{Daneshjou2022-vk, Ktena2023-sl}. Using 3,699 images from the Fitzpatrick 17K\cite{Groh2021-zx} dataset and 656 images from the Stanford Diverse Dermatology Images (DDI) dataset\cite{Daneshjou2022-vk}, we produced 458,920 synthetic images using the latent diffusion model Stable Diffusion\cite{Mostaque2022-uw} fine-tuned with DreamBooth\cite{Ruiz2022-eg}. We then developed and evaluated skin disease classifiers under various experimental settings, including increasing amounts of real data and synthetic data, across image generation strategies, and alongside traditional data augmentation techniques. 

\begin{figure}
  \centering
  \includegraphics[width=16cm]{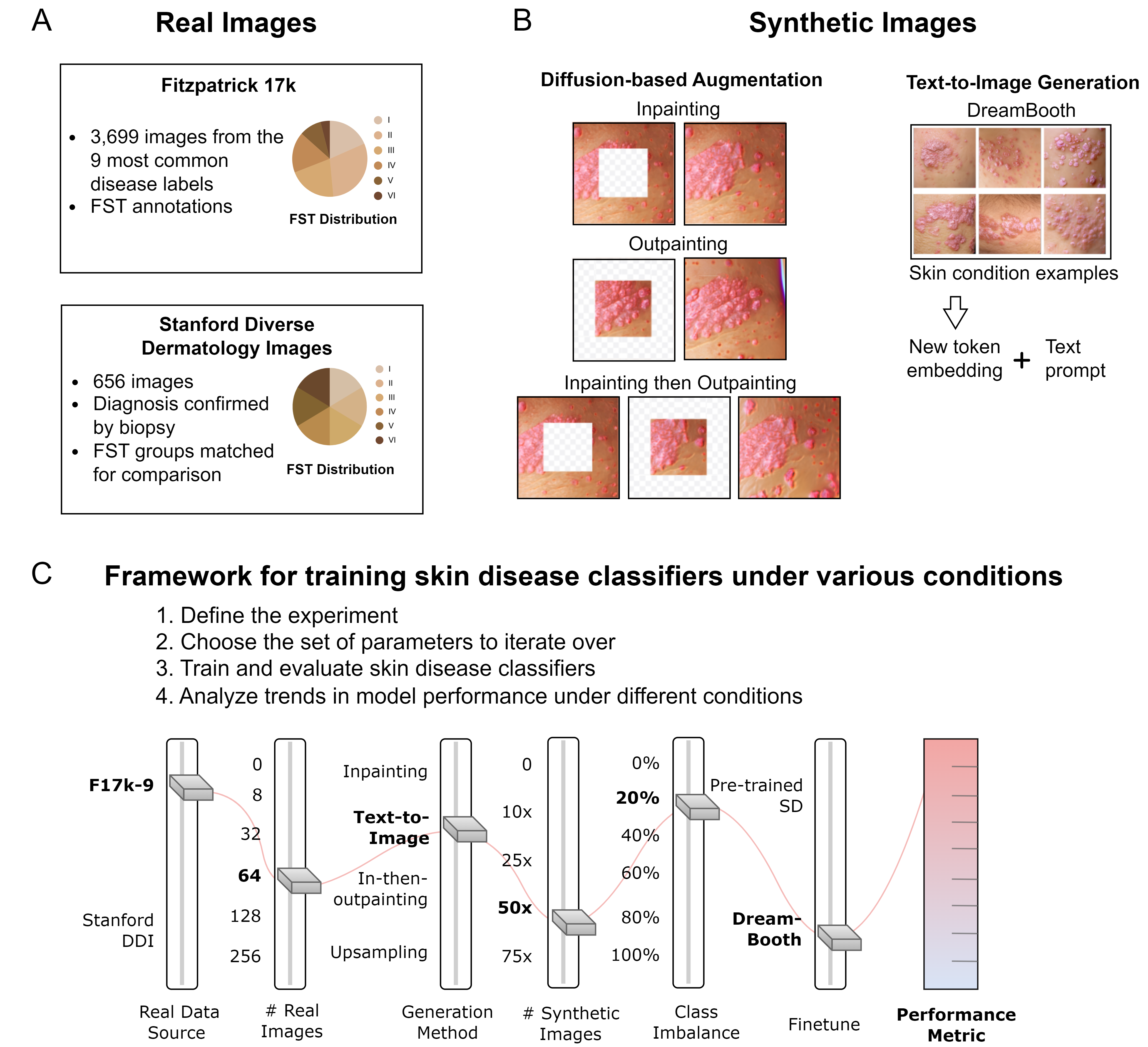}
  \caption{\textbf{A.} Sample sizes, label derivations, and Fitzpatrick skin type (FST) distributions for the datasets used in this study. \textbf{B.} Illustrations of four synthetic generation procedures. Inpainting replaces a central square segment with new generated pixels. Outpainting replaces border regions outside this center square. In-then-outpainting performs these two procedures sequentially to replace every pixel of the image. Text-to-image uses Stable Diffusion fine-tuned using DreamBooth to generate embedding tokens from a specific class. These tokens are  then used to generate new images of that class. \textbf{C.} Data, methodology, and other parameters that affect training and evaluation of skin classifiers. F17k-9 refers to the subset of Fitzpatrick 17k comprising the nine most common skin diseases. }
  \label{fig:fig1}
\end{figure}

\section{Results}
\label{sec:Results}

\subsection{Generating half a million synthetic images across nine disease conditions}
Model training and testing were conducted using images from the nine most common skin conditions in Fitzpatrick 17k: psoriasis (624), squamous cell carcinoma (480), lichen planus (452), basal cell carcinoma (446), allergic contact dermatitis (398), lupus erythematosus (337), sarcoidosis (326), neutrophilic dermatoses (324), and photodermatoses (312), which we term the F17k-9 dataset (Figure\ref{fig:fig1}A). The Stanford DDI dataset is comprised of images of both malignant (171) and benign (485) skin conditions. Both datasets represent the full spectrum of skin tones from Fitzpatrick I-VI (Figure \ref{fig:fig1}). Using text prompts with and without reference images, we generated 458,920 synthetic images; generation methods included inpainting (replacement of central pixels with generated pixels), outpainting (replacement of peripheral pixels with generated pixels), in-then-outpainting (replacement of all pixels by sequential inpainting then outpainting), and text-to-image (de novo generation from text alone) (Figure \ref{fig:fig1}B). A sample of generated images can be found in Figure S-4.

\begin{figure}
  \centering
  \begin{subfigure}[A]{0.55\textwidth}
     \includegraphics[width=\textwidth]{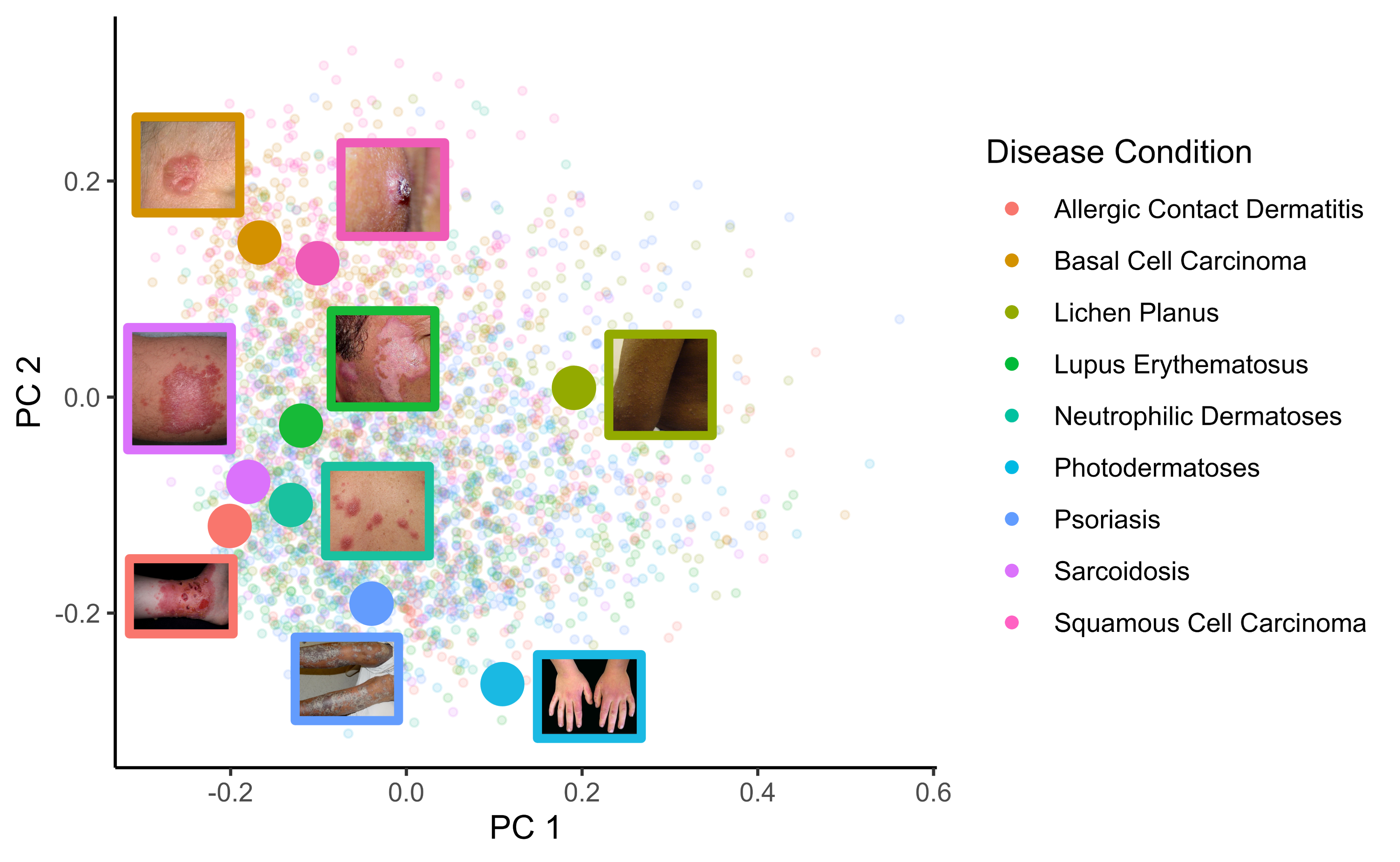}
     \caption{Real images by disease condition}
     \label{fig:fig2A}
 \end{subfigure}
  \begin{subfigure}[B]{0.55\textwidth}
     \includegraphics[width=\textwidth]{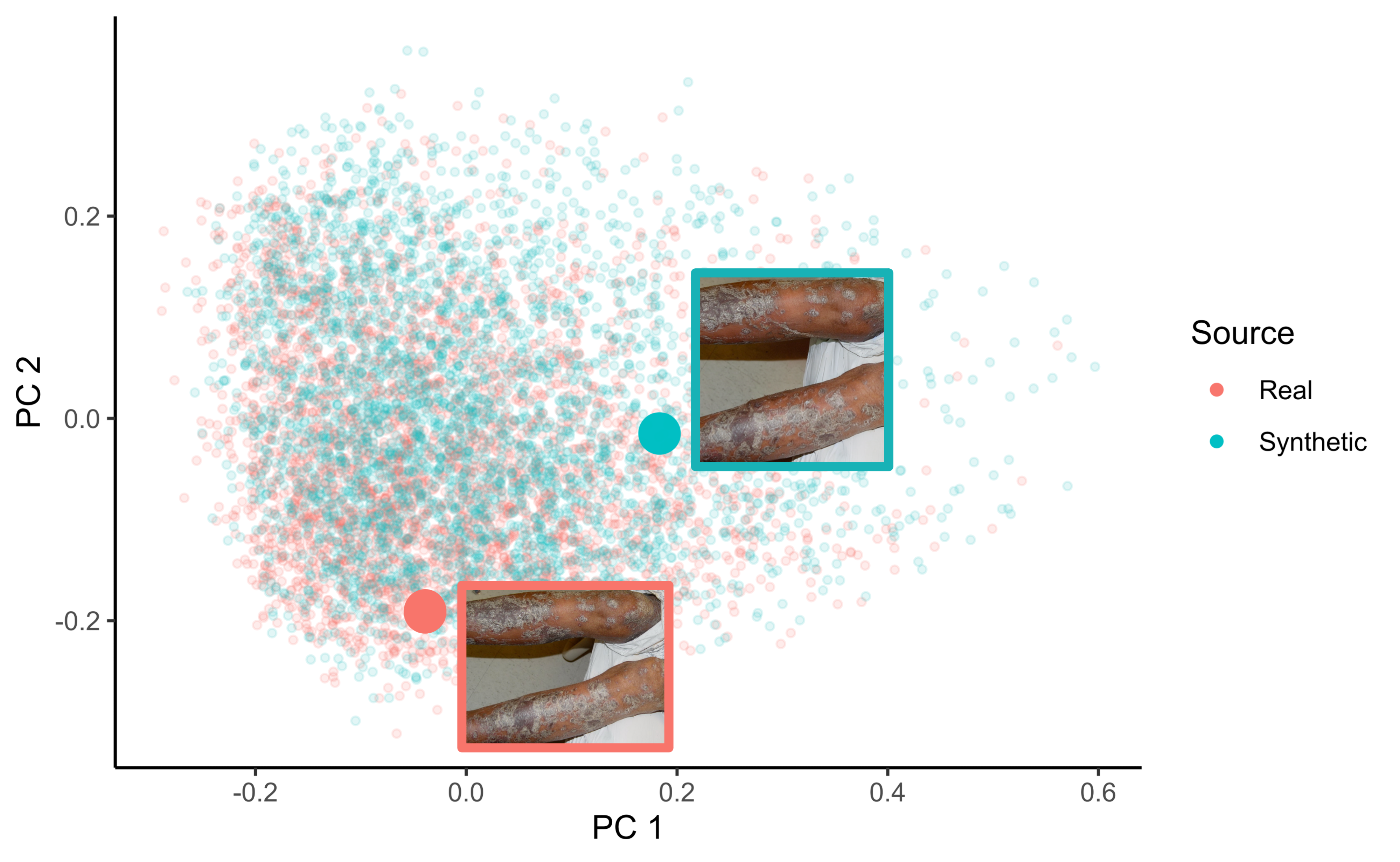}
     \caption{Real versus synthetic images}
     \label{fig:fig2B}
 \end{subfigure}
  \begin{subfigure}[C]{0.8\textwidth}
     \includegraphics[width=\textwidth]{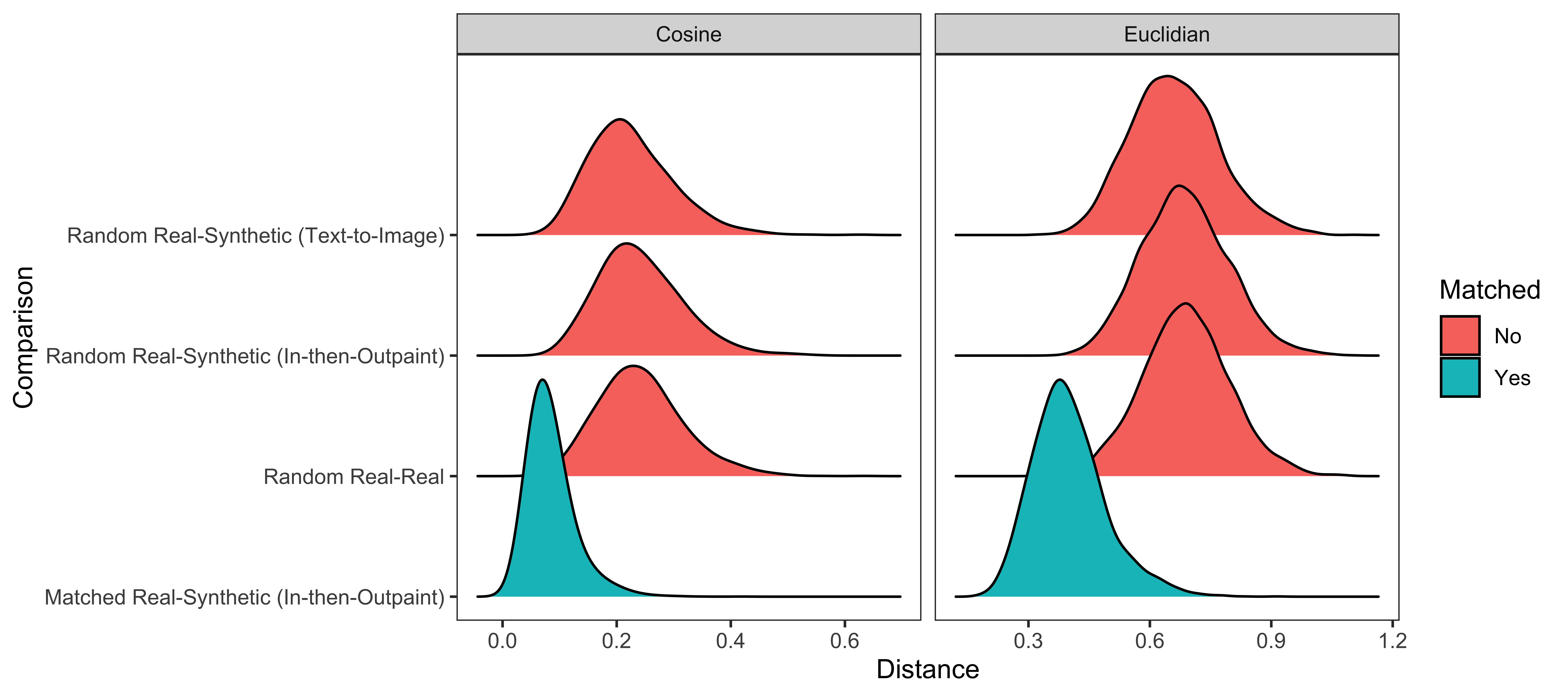}
     \caption{Pairwise distances between image embeddings}
     \label{fig:fig2C}
 \end{subfigure}
  \caption{\textbf{A.} First two principal components of image embeddings for real and synthetic images, colored by disease condition, with overlaid images of randomly selected examples. 
\newline\textbf{B.} Same as A, but colored to indicate real or synthetic data, with an overlaid pair of a synthetic image generated using the in-then-outpaint method alongside its reference image. 
\newline\textbf{C.} Comparison of matched and unmatched pairwise distances between image embeddings of length 512 generated using the image encoder component of the Contrastive Language-Image Pre-Training (CLIP) Vision Transformer (ViT) B-32\cite{Radford2021-hc}. Matched distances between in-then-outpaint images and their respective reference images (blue) were compared to unmatched distances between real and synthetic images and between pairs of real images (red).}
  \label{fig:fig2}
\end{figure}

Image examples are shown overlaid on the first two principal components of image embeddings generated across real and synthetic images (Figure \ref{fig:fig2A}, \ref{fig:fig2B}). Non-linear projections were also produced using uniform manifold approximation and projection\cite{mcinnes2018umap} (Figures S-3). Mean cosine distance between random real-real image pairs (0.242) was similar to that between random real-synthetic image pairs generated using the in-then-outpaint method (0.241, Benjamini-Hochberg [BH] adjusted two-sample t-test p-value: 0.62) and larger than that between these synthetic images matched to their respective reference images (0.0844, BH adjusted two-sample t-test p-value: <0.0001) (Figure \ref{fig:fig2C}).

\subsection{Synthetic data augmentation improves performance in low-data settings and is additive with image transforms}
\begin{figure}
  \centering
  \includegraphics[width=16cm]{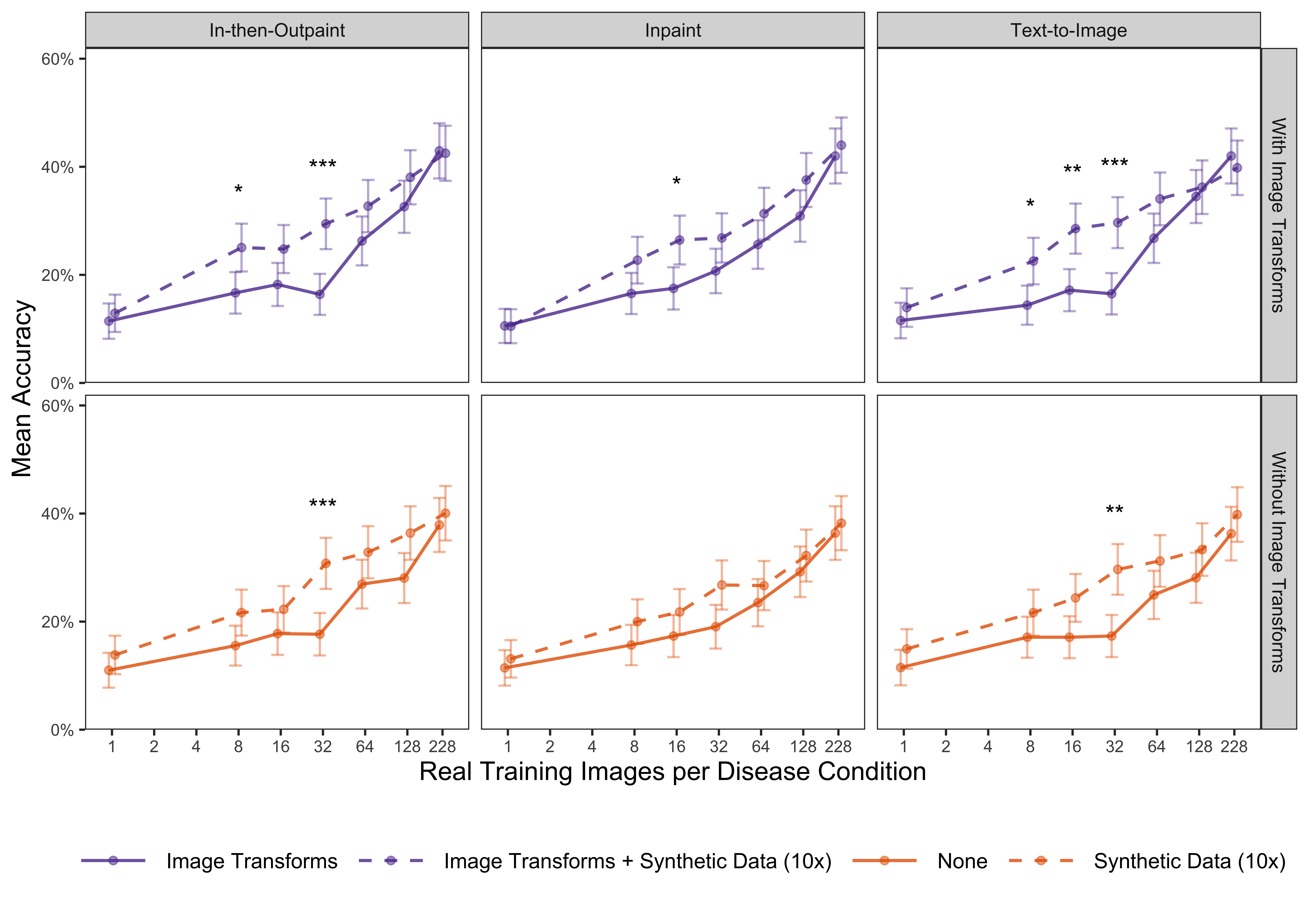}
  \caption{Performance of models trained for nine-way skin classification while varying the number of real images (1 to 228), use of synthetic images (0 or 10 synthetic images per real image), and use of an image transform augmentation utility (yes or no). Model accuracy is averaged across five runs on a balanced, held-out test set of 360 images. Synthetic augmentation was performed using one of three methods: inpaint only, in-then-outpaint, and text-to-image. Error bars represent 95\% confidence intervals computed using the Wald method. Asterisks represent p-values computed using two-sample proportion tests comparing mean accuracy with and without synthetic data at each number of real training images, adjusted using the Benjamini-Hochberg procedure: *p<0.05, **p<0.01, ***p<0.001.}
  \label{fig:fig3}
\end{figure}

We trained and tested skin disease classifiers on a one-of-nine prediction task using the F17k-9 dataset. Two types of training augmentation were evaluated: synthetic augmentations, comprising 10 synthetic images for each real image, or image transforms, comprising random flipping, cropping, rotating, warping, and lighting changes. As expected, classifiers improved with more real images and with image transforms (Figure \ref{fig:fig3}). In addition, synthetic training data improved accuracy of classifiers with and without image transforms. Accuracy gains were highest at 16, 32, or 64 real images per disease class. At 32 real images per disease label and with image transforms, adding synthetic data improved accuracy by up to 13.2\% (BH adjusted p-value = $9.0 \times 10^{-4}$). Accuracy gains diminished at sufficiently large samples of real images; at 228 real images per disease label and with image transforms, accuracy changed by 2.0\% (95\% bootstrapped CI: -1.2\%, 5.3\%) for inpainting, -0.4\% (95\% CI: -3.0\%, 2.0\%) for in-then-outpainting, and -2.2\% for text-to-image generation (95\% CI: -5.5\%, 1.1\%), with none meeting significance thresholds after adjusting for multiple testing.

Adding synthetic data occasionally improved a data-scarce model to the performance of a model trained with more real data. For example, using 32 real images per condition with image transforms and synthetic data (1:10 real:synthetic ratio) from in-then-outpainting performed similarly to using 64 real images with image transforms and without synthetic data (29.4\% vs. 26.3\%, p-value = 0.39) (Figure \ref{fig:fig3}). Performance gains from synthetic data augmentation varied by disease condition and generation method (Figure S-1). Performance increased for all disease conditions except sarcoidosis when augmented using the text-to-image generation method. The greatest performance gains were accrued for prediction of allergic contact dermatitis using the text-to-image generation method.

\subsection{Improvements from synthetic data exhibit dose-response and saturation effects}
\begin{figure}
  \centering
  \includegraphics[width=16cm]{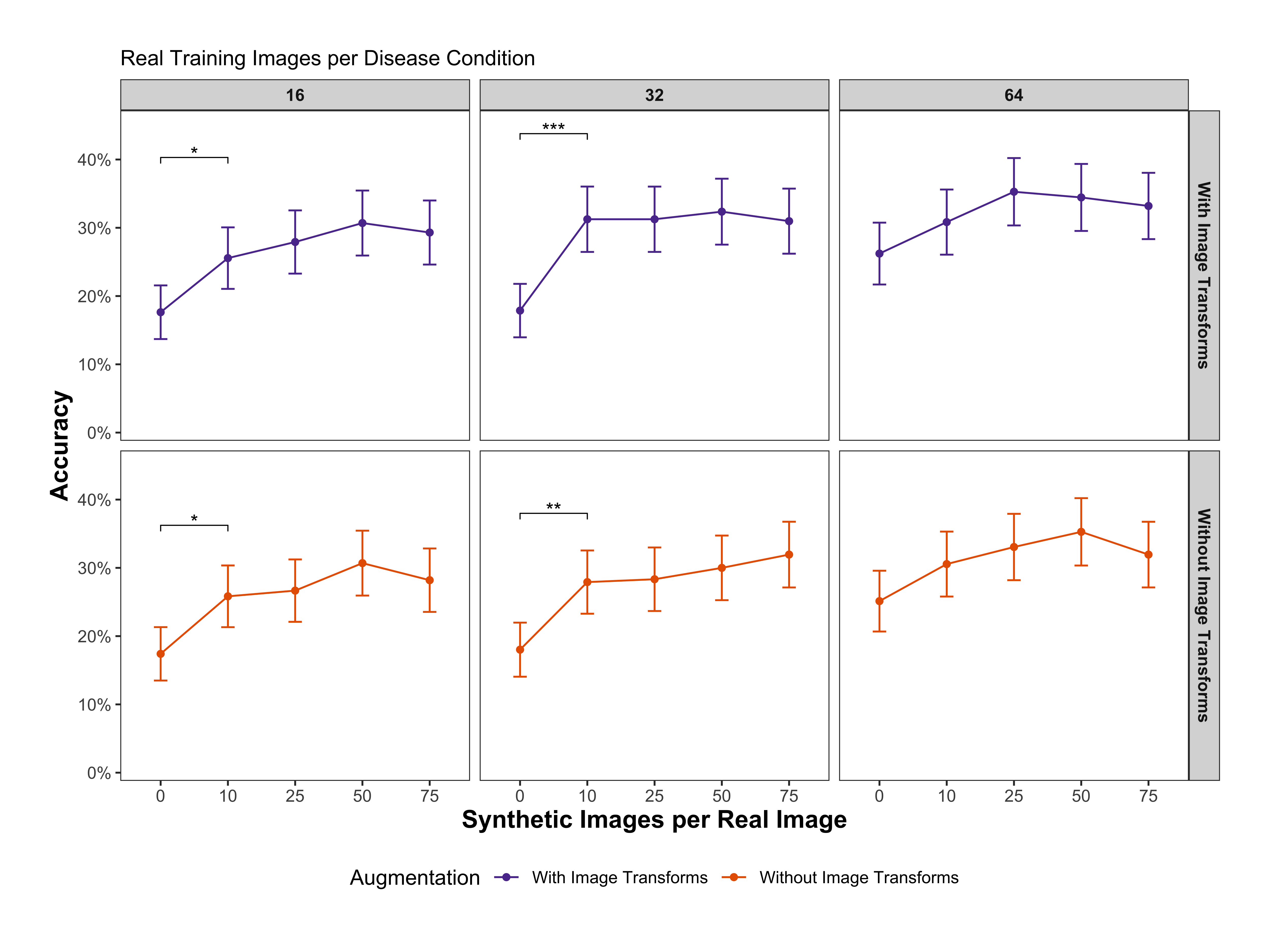}
  \caption{Model performance improvements from augmenting training with 0, 10, 25, 50, or 75 synthetic images per real training image, stratified by number of real images available (16, 32, or 64). Model performance is reported as the mean accuracy across five runs on a held-out test set. Evaluations were conducted with and without image transforms (e.g., flipping, cropping, rotating, and zooming). Synthetic images were produced using the text-to-image procedure DreamBooth. Error bars represent 95\% confidence intervals computed using the Wald method. Asterisks represent p-values computed using two-sample proportion tests and adjusted using the Benjamini-Hochberg procedure: *p<0.05, **p<0.01, ***p<0.001.}
  \label{fig:fig4}
\end{figure}

We trained skin disease classifiers on a one-of-nine prediction task while adding increasing numbers of synthetic images to the training set. Starting with training sets of 16, 32, and 64 real images per disease class, we added 0, 10, 25, 50, or 75 synthetic images per real image; synthetic examples were generated using the text-to-image method due to computational expense (Figure \ref{fig:fig4}). With 16 or 32 real images per class, adding 10 synthetic images per real image led to significant increases in accuracy, up to an absolute increase of 13.3\% (BH adjusted p-value = $5.3 \times 10^{-4}$). Further increases up to 75-fold augmentation sometimes improved performance, but effects were neither monotonic nor statistically significant (Figure \ref{fig:fig4}).

\subsection{Synthetic augmentation improves skin malignancy classifiers across skin types}
\begin{figure}
  \centering
  \includegraphics[width=16.25cm]{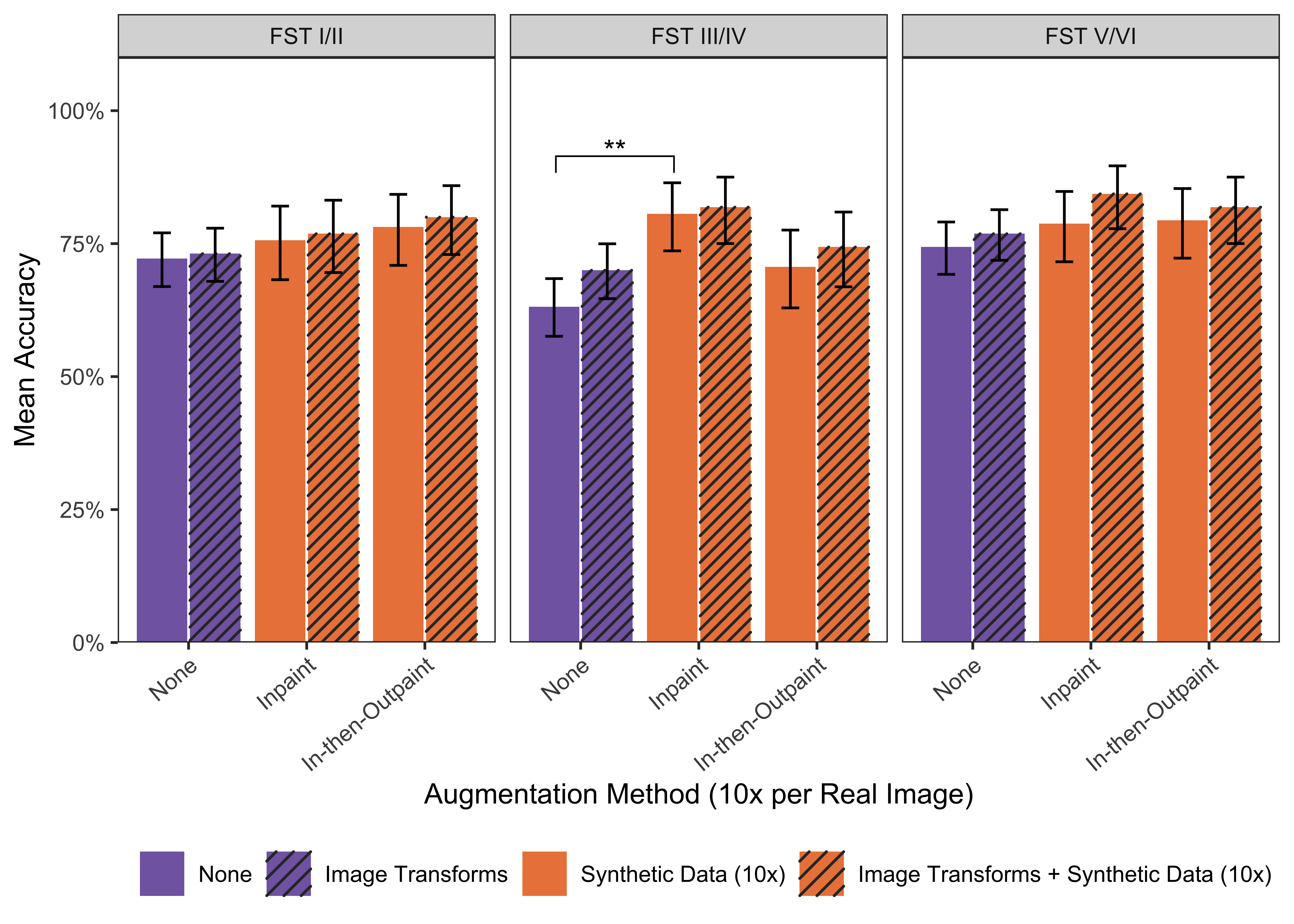}
  \caption{Model performance improvements from augmenting training with 0, 10, 25, 50, or 75 synthetic images per real training image, stratified by number of real images available (16, 32, or 64). Model performance is reported as the mean accuracy across five runs on a held-out test set. Evaluations were conducted with and without image transforms (e.g., flipping, cropping, rotating, and zooming). Synthetic images were produced using the text-to-image procedure DreamBooth. Error bars represent 95\% confidence intervals computed using the Wald method. Asterisks represent p-values computed using two-sample proportion tests and adjusted using the Benjamini-Hochberg procedure: *p<0.05, **p<0.01, ***p<0.001.}
  \label{fig:fig5}
\end{figure}

We trained skin disease classifiers to classify malignant versus benign skin conditions in the Stanford DDI dataset and compared classifier performance between Fitzpatrick Skin Type (FST) labels. Adding synthetic images improved the accuracy of malignancy classification for all FST categories. Notable improvements include 73.1\% to 80.0\% (difference of +6.9\%, 95\% CI: 1.2\%, 13.1\%) in FST I-II using the in-then-outpainting method, 70.0\% to 81.9\% (difference of +11.9\%, 95\% CI: 5.6\%, 17.8\%) in FST III-IV using the inpainting method, and 76.9\% to 84.4\% (difference of +7.5\%, 95\% CI: 2.8\%, 13.8\%) in FST V-VI using the inpainting method (Figure \ref{fig:fig5}). Although adding synthetic images consistently improves accuracy, only one comparison, the addition of synthetic data without image transforms, remained significant following correction for multiple hypothesis testing (31.3\% vs. 17.9\%, BH adjusted p-value = $5.3 \times 10^{-4}$).

\section{Discussion}
\label{sec:Discussion}
In this study, we systematically evaluated the effect of augmenting skin disease classifiers with synthetic images generated by latent diffusion models. In the process, we generated a repository of 458,920 synthetic images, which we will release for public use. We found that synthetic images may improve classifier performance in some data-limited settings with monotonic dose-response behavior, but that improvements saturated at synthetic-to-real data ratios above 10:1. Significant effects were observed across several generative methods and skin conditions, but the primary driver of performance was the number of real images. Prior work has suggested using synthetic images generated from majority groups to supplement groups that are underrepresented (e.g, images of dark skin tones) in training datasets\cite{Rezk2022-ev}. Our work demonstrates that real data is still key and suggests that imbalance in the use of real images versus synthetic images across protected classes could lead to performance differences.

Prior work investigating the use of synthetic data across several medical domains has shown that synthetic images can improve disease and tissue classifiers\cite{Chambon2022-je,Levine2020-bn, Chen2021-fz, Xue2021-my, Giuste2023-gj}, increase model performance and fairness on out of distribution data\cite{Ktena2023-sl, Rajotte2021-pa}, and capture feature distributions for data sharing\cite{Ghalebikesabi2023-bi, Rajotte2021-pa, Ding2023-ok, DuMont_Schutte2021-hi}. Although generative adversarial networks (GANs) have been able to produce photorealistic clinical dermatology images\cite{Ghorbani2019-cr, Rezk2022-ev, Chen2021-fz}, they have not been shown to meaningfully improve performance\cite{Ghorbani2019-cr}. Possible explanations for improved performance of diffusion models include higher image fidelity, greater image diversity, and an increased level of control in the generation process. Diffusion-based methods, which are able to produce high-fidelity and photorealistic images based on text input\cite{Azizi2023-wr}, have produced promising results for skin disease\cite{Akrout2023-rl, Ktena2023-sl}. Our study builds off of preliminary findings that diffusion models can generate data to improve classifier performance for rare disease\cite{Packhauser2022-vd} and across underrepresented groups\cite{Sagers2022-nn}, with experiments that isolate and better understand the impact of synthetic images from diffusion models for natural images in the medical domain.

This study has limitations. First, our analysis was limited to nine skin conditions with several hundred images each. We therefore could not assess whether effects persist at higher sample sizes. Second, we did not assess all factors that may affect synthetic augmentation, e.g. model architecture. Third, the Fitzpatrick Skin Type is a measure of photosensitivity rather than skin tone, though it has been co-opted for assessing skin tone in machine learning applications\cite{Groh2022-qp}. Nevertheless, this scale is not granular enough to capture the full spectrum of human skin tone diversity. Fourth, in the Fitzpatrick 17k dataset, diagnostic labels were not confirmed by biopsy and FST labels were derived retrospectively. Fifth, we cannot confirm that all test data in Fitzpatrick 17k was excluded from pre-training for Stable Diffusion or similar large-scale models; by contrast, images in DDI were released later in November 2022 and required registration. Future work may compare diffusion models head-to-head with previous approaches (e.g., GANs), directly quantify photorealism (e.g., using a Turing test for human clinicians), and investigate the use of text conditioning on anatomic location or disease subtypes.

Recent rapid improvements in generative modeling have prompted significant interest in augmenting medical machine learning classifiers with synthetic images. Our data suggest that synthetic augmentation approaches are worth exploring when sufficient reliable test data is available to assess empirical benefits and harms. While collection of diverse real-world data remains the most important and rate-limiting step for improving skin classification models, achieving parity in representation or performance across hundreds of skin conditions remains challenging, especially across disease subtypes, anatomic locations, and other image characteristics\cite{Combalia2022-dk, Finlayson2021-ep}. The concomitant use of synthetic data, along with traditional methods of re-weighting and upsampling, may act as a force-multiplier to continually improve classification models for healthcare applications.

\section{Methods}
\subsection{Data description}
Skin disease image data are from the Fitzpatrick 17k dataset\cite{Groh2021-zx} and the Stanford DDI dataset12\cite{Daneshjou2022-vk}. Fitzpatrick 17k consists of 16,577 clinical images from two dermatology atlases, including both skin condition labels and FST labels. Although there are 114 skin conditions in the full Fitzpatrick 17k dataset, our analysis utilized only the nine most common conditions with sample sizes sufficient for the proposed experiments; these were selected in order to have at least 228 real images available for each dataset after allocation of validation and testing data. Potentially duplicated, mislabeled, and outlier samples were removed using the SelfClean algorithm\cite{Groger2023-fc}. After filtering, selected conditions totaled 3,699 images (F17k-9). The Stanford DDI data consisted of 656 biopsy-confirmed clinical images representing 570 unique patients from pathology encounters at Stanford Clinics. Additionally, DDI contains FST labels for individuals intentionally matched on diagnostic category, age within 10 years, gender, and date of image within 3 years for comparison between FST I-II and FST V-VI.

\subsection{Synthetic data generation}
Data was synthesized according to three different generation methods: (1) text-to-image, (2) inpainting, and (3) inpainting followed by outpainting. For text-to-image, we generated images by prompting a text-to-image diffusion model with the prompt “An image of \textit{<class-name>}, a skin disease”, where \textit{<class-name>} was replaced by the name of the corresponding disease (e.g., psoriasis). For inpainting, we used a text-to-image inpainting model to mask and replace a square region at the center of the image with width and height equal to half that of the original dimensions (Figures 1-2). Our inpainting model is also conditioned on text; we use the same text prompt as described above. For in-then-outpainting, we first performed inpainting as described above, and then masked and replaced the border of the image, which is an exact complement of the inpainting mask (Figure \ref{fig:fig1}B). As before, we used the same text prompt as input.

All methods utilized Stable Diffusion\cite{Mostaque2022-uw}, a large-scale latent diffusion model, for generating high-resolution synthetic images. The text-to-image method used Stable Diffusion v2.1 and the inpainting-based methods used Stable-Diffusion-Inpainting v1.5. For each of our three generation methods, we considered two variants of the generative model: one in which the pretrained diffusion model was used directly, and one in which the model was fine-tuned on a small number of training images from Fitzpatrick 17k using DreamBooth\cite{Ruiz2022-eg}. Images used for fine-tuning were never included in held-out test sets. 

\subsection{Model training}
Our experiments utilized the Efficient-Net V2-M image classification model pretrained on ImageNet\cite{Tan2021-zv}. In all experiments, training was performed for 30 epochs; model training was stopped when validation accuracy did not improve in 3 consecutive epochs. The data was split into training and validation sets (Table S-1) and a held-out test set was used for final evaluation. Synthetic data was added to the training set only and was always produced independently of images in held-out test sets. Some experiments involved image transforms including random cropping, horizontal flips, warping, and lighting adjustments from the aug\_transforms utility in fast.ai\cite{Howard2020-vk}, applied to both real and synthetic images. 

\subsection{Model evaluation}
We created separate training sets comprising 1, 16, 32, 64, 128, and 228 images per disease condition (totaling 9, 144, 288, 576, 1152, and 2052 real images) in F17k-9. For each training set, we trained skin disease classifiers in four ways: (1) using only the real images without image transforms (2) using only real images with image transforms, (3) using real images, 10 synthetic images per real image, and no image transforms, and (4) using real images, 10 synthetic images per real image, and image augmentation. Validation loss was calculated on a set of 40 images per disease label, totaling 360 images. After training, mean classification accuracy was calculated on a held-out test set of 40 images per disease label, totaling 360 images (Figure \ref{fig:fig3}). To account for stochasticity in model training, we report mean accuracy over five runs for each set of training criteria.

\subsection{Dose-response and saturation experiments}
We measured the effect of adding increasing amounts of synthetic data to the training set of skin disease classifiers. The training set of each model comprised either 16, 32 or 64 real images per skin condition. Then, to each training set, we added either 10, 25, 50, or 75 synthetic images per real image, generated with the text-to-image method. Models were trained with and without image transforms. We evaluated mean classification accuracy for each model on the same held-out test set comprising 40 real images per skin condition.

\subsection{Class imbalance experiments}
We also simulated a commonly cited use case for synthetic data: mitigating class imbalance for under-represented labels in a training dataset\cite{Chen2021-fz, Rajotte2021-pa}. Using the F17k-9 dataset, we selected 200 examples from each disease condition, and intentionally downsampled one of these to 25, 50, or 100 images during model training. For each imbalanced dataset, we trained a skin disease classifier and evaluated its accuracy for the rare disease class on a held-out test set. We then retrained the models after adding different numbers of synthetic images to the rare disease class. We added 1, 3, or 7 synthetic images per real image of the rare disease class to achieve parity with the non-downsampled conditions. As a control, we also tested an upsampling strategy of simple duplication for the rare disease images. We reported the mean classification accuracy of each model on the rare disease classes separately as well as in aggregate (Figure S-2).

\subsection{Skin malignancy classification by skin tone}
We measured the effect of including synthetic data when training a malignant vs. benign skin-disease classifier on FST-balanced DDI data. We removed 32 images per FST category (I/II, III/IV, V/VI) to create a 96-image held-out test set and then fine-tuned a Stable Diffusion model for synthetic images generation using the remaining 560 images. We trained models to classify images as either malignant or benign while varying the following conditions: use of synthetic data (10 synthetic images per real image), use of image transforms, and generation method (inpaint or in-then-outpaint). For each combination, we reported the mean classification accuracy of models across FST categories.

\section*{Acknowledgments}
This project was funded in part by the National Heart Lung and Blood Institute (NHLBI) award K01HL138259 and by the National Library of Medicine (NLM) award 2T15LM007092-31.

\printbibliography

\end{document}